# Parking Space Ground Truth Test Automation by Artificial Intelligence Using Convolutional Neural Networks

Tony Rohe
*Cross-Domain Computing Solutions*
*Data Acquisition Technology*
Robert Bosch GmbH
Leonberg, Germany
tony.rohe@de.bosch.com

Martin Margreiter
*Mobility Partners*
Munich, Germany
martin@mobility-partners.com
https://orcid.org/0000-0002-0428-0914

Markus Mörtl
*Laboratory for Product Development and Lightweight Design*
*TUM School of Engineering and Design*
Technical University of Munich
Munich, Germany
markus.moertl@tum.de
https://orcid.org/0000-0002-4122-1307

*Abstract*—This research is part of a study of a real-time, cloud-based on-street parking service using crowd-sourced in-vehicle fleet data. The service provides real-time information about available parking spots by classifying crowd-sourced detections observed via ultrasonic sensors. The goal of this research is to optimize the current parking service quality by analyzing the automation of the existing test process for ground truth tests. Therefore, methods from the field of machine learning, especially image pattern recognition, are applied to enrich the database and substitute human engineering work in major areas of the analysis process. After an introduction into the related areas of machine learning, this paper explains the methods and implementations made to achieve a high level of automation, applying convolutional neural networks. Finally, predefined metrics present the performance level achieved, showing a time reduction of human resources up to 99.58 %. The overall improvements are discussed, summarized, and followed by an outlook for future development and potential application of the analysis automation tool.

## I. INTRODUCTION

The general trend towards growing cities and an increasing demand for individual and flexible mobility significantly increases the problem of parking. The increasing pressure on parking spaces results from historically grown road networks, increasing level of motorization and commuting, non-local demand through cultural attraction and an increasing demand of infrastructure for non-motorized users (bicycle paths) [1]. Furthermore, the market tends to Sport Utility Vehicles (SUVs) and larger cars, which additionally increases parking space requirements [2]. An external partner is developing a real-time parking service, to tackle the parking problem in city centers. While the aim is to provide information about parking costs, special parking spaces (like disabled parking spaces) and the prediction of areas with a comfortable parking situation, the core service is to provide real-time information on free and available parking spaces.

The underlying idea is the community approach: every car providing real-time information also benefits from the final processed information. Ultrasonic sensors in the transmitting vehicles detect raw spaces (simple gaps) along the street enclosed with objects like cars and trees. This information (called detections) is sent via the OEM infrastructure to the service cloud backend. There, the incoming detections are processed and labeled as "parking" or "non-parking". Non-parking detections can be crossings, entrances and restricted areas. The detections labeled as "parking" are called parking spots. Finally, the availability of parking spots and their location is provided as a service to the receiver vehicles. More information on the parking assistance tools using in-vehicle data is described in studies like [3], [4], [5] and [6].

In order to evaluate the current service quality as well as to support the continuous development, system tests are conducted capturing the test drive with a video camera. These videos are evaluated afterwards by test engineers to create a reference label which can be used to score the performance of the parking service. These tests are called ground truth tests, as they test the system with real-world data. Although these tests include all possible parking situations and variances and directly measure the main Key Performance Indicator (KPI) of the service quality, the False Positive (FP)-rate, they come with big disadvantages – immense costs and human efforts. As a rule of thumb, test analysis exceeds the time of the test drive by a factor of 40, due to the requirement of licensed software, inconvenient data handling and finally the manual repetitive task of finding the content of hundreds of excel rows in low performing video visualization to label individual parking sequences.

The aim of this research is to eliminate these disadvantages. The test analysis should be fully automated, fast processing, reproducible, with a maximal use of the data available but at the same time, and most importantly, with as few errors as possible to allow a valid measure of the KPIs. The concept introduced in this research is to use a trained Convolutional Neural Network (CNN) to classify the images of the reference video and to conclude based on this classification on a label for each detection. Therefore, an automation tool and an analysis tool are developed, implemented and evaluated in new ground truth tests. The aim of this paper can be summarized to a single question:

*"How are savings in time and human resources in ground truth test analysis possible by using machine learning methods?"*

## II. METHODOLOGY

This chapter introduces the main methods used to achieve a high level of test analysis automation by applying machine learning with CNN.

### A. Machine Learning

A condensed but still vivid definition of machine learning is given by Tom Mitchell:

> *"A computer program is said to learn from experience E with respect to some task T and some performance measure P, if its performance on T, as measured by P, improves with experience E."* Tom Mitchell, 1997. [7]

Based on experience (training data) a model is able to perform a task with a certain performance. Performance improves with experience (a more comprehensive training data set). This illustrates why one of the major problems in Machine Learning (ML) is to acquire good and sufficient training data. This means that ideally all problem variations, which can occur in the task, are represented in the data set. Furthermore, the data must correlate to each other. Correlation means that there is a relationship or association between the input and the output data. This concept can be applied in computers using artificial neural networks, or for complex problems and with a more performant topology Deep Neural Networks (DNN). More information on ML, neural networks, deep neural networks and how to train these algorithms on a specific problem can be found in [8] and [9].

### B. Convolutional Neural Networks (CNN)

CNNs are especially well suited for grid-like input data like one dimensional time-series data or two-dimensional image data. These networks employ a mathematical function called convolution which is decisive for the naming. Goodfellow et al. [8] puts it clear saying:

> *"Convolutional networks are simply neural networks that use convolution in place of general matrix multiplication in at least one of their layers."* [8]

The mathematical concept of convolution applied on image pattern recognition tasks in CNNs is implemented as shown in Figure 1. A kernel matrix, also called filter-matrix slides over the image, in this case restricted by the fact that the kernel lies entirely within the image. This operation is referred to as "valid" convolution. Other convolution types are referred to as "full" and "same" convolution. Starting in the top left corner, corresponding to their location, a is multiplied by w, b is multiplied by x and so on until finally the top left value in the output matrix is calculated by the sum of the multiplications. Because the kernel matrix is a 2x2 matrix in this example, the 3x4 input matrix is converted to a 2x3-output matrix. This operation, if applied in a larger neural network architecture, is referred to as a convolution layer. Sequential convolutional layers are able to extract more and more complex shapes, structures and characteristics of an image. These features can finally be used to conclude on a probability estimation for class membership, object detection or semantic segmentation. Further information on CNNs is given by Goodfellow et al. [8].

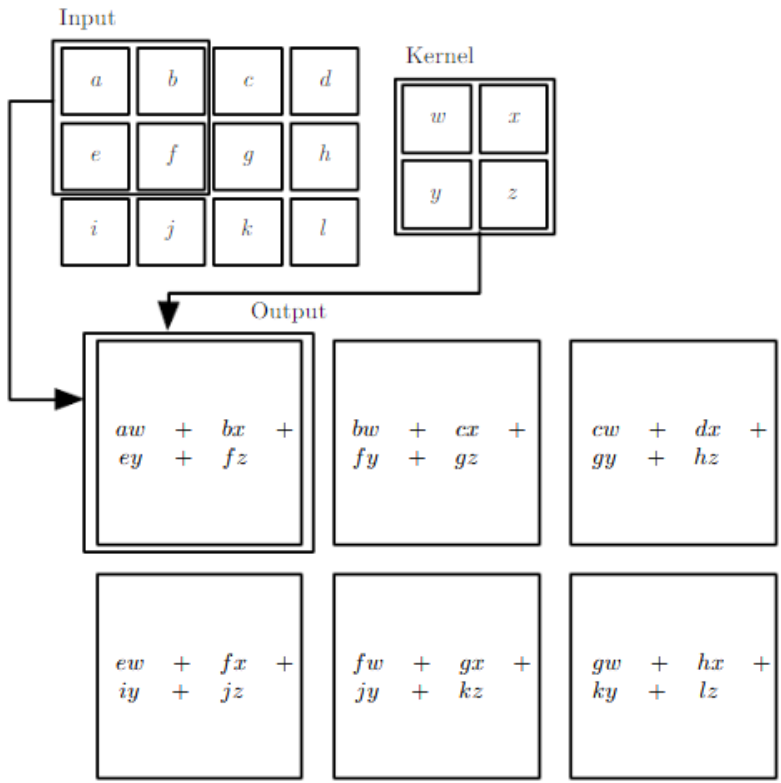

Fig. 1. Example of "valid" convolution of an input matrix with the kernel matrix creating an output matrix [8].

## C. Test

As a substance of matter, testing needs a system under test and therefore happens at the end of each development cycle. As development projects tend to run late in their tight schedule, acceleration and therefore automation of as many test processes as possible is necessary. Several models describe the role and responsibility during the development cycle. The common model currently applied in the automotive industry is the V-model [10]. Generally, the project is decomposed in the downward diagonal into the definition of its subsystems and components and on the upward diagonal the verification ensures the implementation as required from unit test level via subsystem level upwards until final system and end-to-end tests. As this research focusses on the right upward diagonal, verification and validation are defined in ISO 24765 [11]. These terms can be described as follows:

- **Verification** regards the capabilities of the initial concept design and evaluates if these requirements are completely implemented correctly – In short: "Are we building the system right?"

- **Validation** considers the final application and examines if the correct set of requirements is established – In short: "Are we building the right system?"

Test is one of the main methods for verification and validation to ensure the developed functionality matches the requirements. The process of testing is understood as "all activities for planning and controlling, analysis and design, realization and implementation, evaluation and reporting as well as concluding and finalizing all test activities necessary in a project." [12]. It can be seen as the main tool, which can be applied for verification as well as validation, dependent on the type of implementation in the overall development process.

## D. Metrics

Several general metrics can be defined to evaluate the quality of a classification algorithm. They are defined and calculated based on quantitative data and are used to compare and track performances, development progress and the achieved level of predefined targets. A few major metrics for binary classification are defined below:

- **True Positive (TP)**: Correctly classified as positive (E.g. a parking spot classified as parking).

- **False Positive (FP)**: Falsely classified as positive (E.g. a garage exit classified as parking).

- **True Negative (TN)**: Correctly classified as negative (E.g. a garage exit classified as non-parking).

- **False Negative (FN)**: Falsely classified as negative (E.g. a parking spot classified as non-parking)

- **Accuracy**: Portion of the classified data being correctly classified $\left(\frac{TP+TN}{TP+TN+FP+FN}\right)$

- **Precision**: Portion of positive classified data, which is correctly classified as positive $\left(\frac{TP}{TP+FP}\right)$. How often are positive classifications correct? → Focus on low false positive rate.

- **Recall**: Portion of positive data which is correctly classified as positive $\left(\frac{TP}{TP+FN}\right)$. How often is a positive output correctly classified as positive? → Focus on low false negative rate.

- **F1**: Combination (harmonic mean) of precision and recall. A good (close to 1) F1 score proves a low FP as well as a low FN $\left(\frac{2 \cdot Precision \cdot Recall}{Precision+Recall}\right)$.

Precision and recall can also be defined focusing on negative classification to calculate an F1 score for non-parking by exchanging "P" and "N". Both F1 scores (for parking and non-parking) can be averaged for a final metric, simplifying the comparison of classification performances.

## III. GROUND TRUTH TEST ANALYSIS AUTOMATION (GTA)-TOOL

In order to achieve the aim of this research of a time and cost reduction in test analysis, the methods explained in the previous chapter are applied in the context of ground truth testing and implemented. This chapter explains in detail, how the vision of a more automated analysis of ground truth tests can be achieved by use of convolutional neural networks. This is done developing a Ground Truth Test Analysis Automation (GTA)-tool [13].

## A. GTA Architecture

The architecture of the GTA-tool shows four key components and several submodules shown in Figure 2. First, a main script provides a console-based user manual that guides the tester through the whole process. The second key component is the data preparation tool processing the large amount of test drive data with corresponding filter submodules for the ultrasonic system via CAN and the near range camera (NRC) system. The third core module, the automation tool, selects the images corresponding to the detections of the test drive and hands it over to its sub module, the CNN, which stores the results of its classification for the last step. Finally, the fourth component, the analysis tool, allows an efficient and easy access to the test data and analysis results. Its first sub module, called decision maker, concludes on the final label for each detection based on the CNN classification. The second sub module, the accuracy tool, exports the key metrics calculated in a test report file.

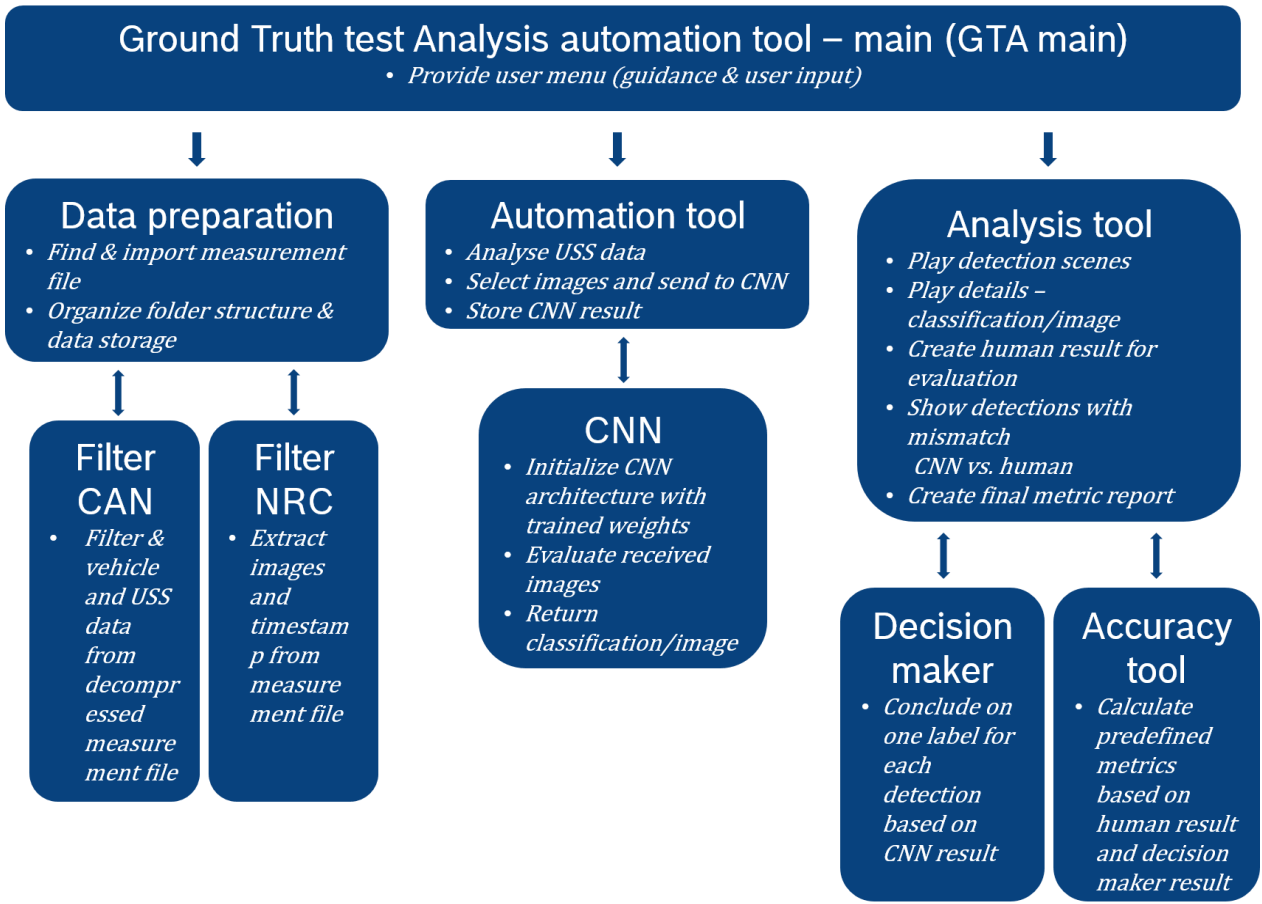

Fig. 2. Architecture of the GTA-tool.

*B. GTA Main*

The main module combines the remaining three key components to one final tool. It makes it easy to start as a single executable or, initially, as a Python script. Afterwards, it guides the user through the multiple options and asks for needed input data like the work path to the measurement files and the current measurement file name(s) to be processed. Advanced users have the possibility to execute the GTA-tool via console on multiple test files providing their paths as optional arguments.

*C. Data Preparation*

After a decompression step, where the compressed measurement files are unpacked, the data preparation tool's only task is to write the necessary information to JSON-files for easy and fast access. The data preparation module mainly accesses preexisting python scripts for loading and filtering data from CAN traces. The output of the data preparation task is the images of the video captured by the NRC and two JSON files. One containing the timestamps of the images and the other containing information about the car odometry and the ultrasonic system (USS) detections.

*D. Automation Tool*

The automation tool takes the role of the test engineer in the old test analysis process. It finds the detections in the video, classifies each image using the CNN, and enables the decision maker to label the detection ID finally with a tag as parking or non-parking.

Therefore, we have to link the local progression of raw space detections, identified with USS chronologically during the test drive with the time sequence of images in the video. Based on the timestamp of the USS detection, the time point of the end of the raw space can be calculated. The relation of time ($\Delta t$), longitudinal distance ($\Delta s$) and the continuous velocity ($v(t)$) of the test vehicle is shown in equation 1.

$$\Delta s = \int_{t_0}^{t_{end}} v(t)dt \quad (1)$$

The USS provides five main measurements for this application. The time point of detection ($t_{det}$), the longitudinal distance to the end of the parking spot at $t_{det}$ called $PS_{Xpos}$, the length of the raw space ($l_{det}$), the velocity for every time point $v(t)$ and an identification number ($ID_{det}$). This measurement data (visualized in Figure 3) is now used to calculate the start ($t_0$) and end time point ($t_{end}$) of the raw space using their relation shown in equation 1.

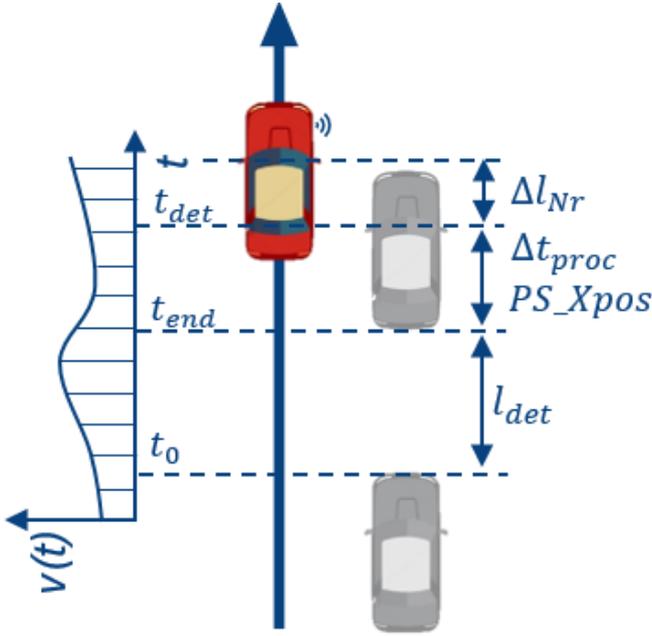

Fig. 3. Visualization of USS measurements taken for each detection.

*E. CNN*

Setting up and training a new CNN from scratch would not have been possible in the scope of this research, essentially as collecting and labeling more than one hundred thousand images needed for training would have taken the majority of the time available. Fortunately, a well-suited trained neural net for a varying but in terms of classification task similar application is found within the research project and implemented. As the optimization of the CNN is not the focus point, only a few characteristics of the CNN are given:

- Framework: Pytorch
- Topology: ResNet (152)
- Output layer: Softmax
- Output classes: 4 (car, construction site, non-parking and parking)
- Training mode: supervised learning
- Training data: >150,000 labeled images

The CNN is implemented in Python 3.7 using the Pytorch framework. The incoming image from the NRC is cropped and interpolated to suit this input requirement. Afterwards the values of all image pixels are fed into the model and evaluated as explained in section "Methodology" above.

The final output layer (Softmax) consists of four nodes representing the classes car, construction site, non-parking and parking. Each image leads to one value for each node, representing the probability that the input image belongs to this specific class. These four values, the output of the CNN, are returned to the automation tool, where all classification results are stored in a three-dimensional array.

The first dimension is the detection ID, making each raw space identifiable. The second dimension is the image ID or image count. It states on which position the image is in the detection. The third dimension contains information for each image. Image count, four values regarding the four label classes (car, construction, non-parking and parking), the image name (similar to the image count but starting with 0 at the beginning of the test drive) and a length weight. This length weight is needed in the successive decision maker. It represents the distance the image is valid for or in other words the distance between the midpoint of the current and the midpoint of the following image. It is calculated knowing the temporal distance between each image due to the NRC timestamp and the current velocity. This distance value is only numerically valid when driving straight ahead. The geometrical error during turning maneuvers is negligible, as parking spots are usually not in sharp corners and the geometrical error in slight curves is negligible and within the measurement accuracy of the velocity sensor.

## F. Analysis Tool

The third module calls the decision maker to conclude on a final label for each detection. It further focuses on the visualization of the results achieved and provides an efficient generation of human results as a reference for the further development of the GTA-tool.

After the generation of the human result, or applied on the parking service, after the import of the system under test result, a test report can be generated automatically showing main service quality KPIs and further predefined metrics.

### 1) Decision Maker

The decision maker draws conclusions based on the classification result of the CNN for each detection on a final label. It creates the final "answer" of the automation tool. This label is created in two different ways, one for standard detections and a separate treatment is implemented for special cases. The pool of possible answers consisted initially of only two options. 0 for non-parking, and 1 for parking. The experience gained with the discovery of the special cases explained below extends the range of possible answers to five. 5 for cross parking spaces and 0.4 and 0.6 for non-parking and parking with low confidence respectively complete the list.

The initial concept intended to conclude only on a normalized weighted average of all classification results of one detection. The weights lower the importance of the images at the sides of the raw space. These images sometimes belong to the images of the cars enclosing the parking space, because the image selection is based on measurement data with a non-zero measurement inaccuracy. The weights reduce the bad influence of these outlier images by shifting the focus to the inner images. This weight function was initially in Gaussian shape with the peak located in the middle of the raw space, but early results showed that a simpler function is sufficient. The first and the last 20 % of the images of a raw space are weighted with only 20 %, while the remaining 80 % is weighted with 100 %. The normalized weighted average for each detection and *class(j)* is calculated according to equation 2, with *wi* being the weight per image, *resCNN,i,j* the result of the CNN of each image *(i)* and each class *(j)* and *nclass* and *nimg* the number of classes (currently four) and the number of images of each detection respectively.

$$res_j = \frac{w_i res_{CNN,i,j}}{\sum_{j=1}^{n_{class}} \sum_{i=1}^{n_{img}} w_i res_{CNN,i,j}} \qquad (2)$$

The class car and non-parking are combined to one class, as the backend labels only as parking and non-parking. In a final step the class with the relative majority is chosen as the final label.

After evaluation of the initial results, several exceptional situations were identified in which the standard application leads to a wrong result. The major exception case, referred to as error case one, is the detection of a small parking spot in a large raw space. The majority of the raw space is non-parking, while a sufficient area for parking remains at the end of the detection. If more than 50 % of the raw space belongs to a non-parking area, the weighted average will result in non-parking, while a car could still park in this raw space.

The solution to this problem is to assign a length weight for each image. Each raw space labeled as non-parking from the weighted average algorithm is checked afterwards, if the sum of the length weights of a continuous sequence of images featuring highest CNN result in the class parking exceeds the length 3.5 m. This length is the project definition of the minimal length of a parking spot. An even more complex case, referred to as error case two, considers the possibility of an ill-posed CNN classification on top of the detection of a small parking spot in a large raw space. Several solution approaches were discussed, and a threshold for the number of consecutive ill-posed images is implemented. An empirical analysis showed good results for a threshold of 2.

Another special case, referred to as error case three, is the inverse problem of the ones discussed above. Too small parking areas in larger detections should be labeled as non-parking. If the raw space is larger than 3.5 m but only 60 % of it is a valid parking area, the weighted average would result in parking, while no car is able to park there. Although the problem is similar, the length weight solution can not be applied here, as this error occurs usually for raw spaces with less than 5m total length and therefore they contain significantly fewer images, and the impact of outlier images is higher.

The solution applied can be described as a length confidence switch. Multiplying the weighted average of the parking class with the length of the raw space has to be higher than a threshold set by experience. A threshold of 3 showed good results and is currently implemented. This means as an example, that a raw space of length 4m has to be classified with at least 75 % (normalized weighted average) as parking to be labeled as parking. All the above application cases are tailored for parking spots, oriented parallel to the street.

A fourth special case is the detection and classification of cross parking spots. These have a longitudinal spacing (raw space length) between 2.1m and 3.5m. Currently these parking spots are detected and classified but categorized in their own class and not further considered, as the parking service focuses on parallel parking spots for the time being.

*2) Low Confidence Switch*

Already during the early stage of this work, the accuracy was prioritized over the level of automation, as the label of the GTA tool is used as a reference to evaluate the main KPIs or the parking service. Therefore, from the initial concept onward, an option to review the result of the decision maker with an experienced test engineer was intended and is implemented in the form of a low confidence switch. This switch sets a threshold, which has to be exceeded for full automation. Each label falling below this threshold, is tagged with a low confidence flag and is recommended to be reviewed before the end of the test analysis.

As an example, if a detection is analyzed by the decision maker and the normalized weighted average of all images belonging to these detections is 5 % car, 2 % construction site, 15 % non-parking and 78 % parking, this detection has a confidence of 78 %. If the low confidence switch threshold is below 78 % the detections are finally labeled as parking, if the threshold is 78 % or higher, it is flagged for manual assessment by the test engineer afterwards. The low confidence flag is implemented as a modification of the conclusion of the decision maker. 0 for non-parking is changed to 0.4 and 1 for parking is changed to 0.6. The analysis tool is able to filter these detections and allows the user to view only these detection sequences to assign the final reference label. The threshold for the standard application is set to 0.7, which means that all normalized weighted averages below 70 % are recommended to be reviewed. This threshold can be adjusted to optimize the tool for higher accuracy or shorter analysis time (see Figure 5 in the next chapter).

*3) View Options*

Several view options are implemented. The main improvement compared to the previous analysis process is the automated focus on the detections. The user is able to play only the detection sequences, while the tool skips the video images in between. This leads to a significant time reduction as well as increasing the focus of the test engineer on the decisive images.

*4) Accuracy Tool*

The metrics, explained in "Methodology" section above, are applied to measure the performance of the classification of the GTA-tool in various categories. The accuracy tool calculates these metrics based on a human reference result and the label of the decision maker. These metrics are shown in the console and saved as a test report in the results sub folder.

## IV. Results

The GTA-tool is designed to automate the analysis part of ground truth tests. It is developed based on an initial test drive. In the scope of this thesis further test drives were executed and the performance of these test data is evaluated in the following subchapter "Main Results". Chapter "Low Confidence Switch Parameter Study" shows the results of a parameter study, varying the sensibility of the low confidence switch introduced in the section "Decision Maker" above.

*A. Main Results*

Based on the findings on the initial data set, several test drives are analyzed to gain statistically more meaningful and valid results. A total of six additional test drives were performed following the guideline of ground truth testing, with a total of 304 detections. The complete test evaluation is presented in Figure 4. Columns one to six show the metrics in the format explained above. Column seven calculates the main metrics based on the sums of the detections in the upper third, representing the overall resulting average performance. Starting with the key metric in the bottom line, the overall performance (F1 score average) reached 82.8 % and 84.7 % considering the low confidence method by almost 10 % of the detections, which confirms the limited impact of the low confidence switch on overall performance.

A deeper analysis of column seven unveils the root cause by an exceptional low performance in classifying only valid parking areas as parking, visualized by the precision metric for positive detections. While the precision for non-parking detection achieves more than 97 %, the precision for parking detections is 62.5 % and 64.7 % depending on the consideration of the low confidence method. The recall values resulting from the six test drives analyzed are located around 90 %. A significant improvement in the overall performance can be achieved by improving the precision of classifying parking detections and therefore reducing the FP rate. Focusing on reducing the FP rate, 24 of 304 detections were FP. A deeper analysis of these 24 cases using the analysis tool of the GTA-tool resulted in four areas for improvement.

More than 50 % of FP cases are related to street signs. The CNN applied is able to relate structures and therefore signs to classes, but no explicit sign recognition is implemented. Doing this would bring significant additional information to enhance the decision-making process. More than a quarter of the FP cases are theoretically possible to avoid, improving the current CNN topology implemented with additional training data. The third improvement area relates to a major problem using CNN in general. If the information needed to classify an image is not available in the picture, no training data set can lead to a correct answer. The fourth action targets a new unseen failure case. The USS reported a raw space just in front of a traffic signal. Due to the waiting time in front of the traffic signal, approximately 900 images are identical, showing a small parking area of 0.3 m within the non-parking detection. The weighted average concluded therefore wrongly with 92 % on parking.

Summarizing the findings from the main result data set, the classification performance is assessed using the selected metrics. Several areas for potential improvement are identified and analyzed. A few cases with weak classification can be tackled with an extended training data set while the largest potential is shown by including a road sign recognition function.

| measurement file name: | test drive 1 | test drive 2 | test drive 3 | test drive 4 | test drive 5 | test drive 6 | sum / average |
|---|---|---|---|---|---|---|---|
| true positive (tp) | 8 | 7 | 16 | 0 | 7 | 2 | 40 |
| false positive (fp) | 5 | 4 | 5 | 5 | 2 | 3 | 24 |
| true negative (tn) | 43 | 33 | 15 | 7 | 35 | 38 | 171 |
| false negative (fn) | 2 | 2 | 0 | 0 | 0 | 1 | 5 |
| true positive low confidence (tp_lc) | 3 | 0 | 0 | 0 | 0 | 1 | 4 |
| true negative low confidence (tn_lc) | 16 | 7 | 10 | 7 | 8 | 12 | 60 |
| total | 77 | 53 | 46 | 19 | 52 | 57 | 304 |
| **without consideration of low confidence** | | | | | | | |
| accuracy | 87,9% | 87,0% | 86,1% | 58,3% | 95,5% | 90,9% | 87,9% |
| *focus on positive detection* | | | | | | | |
| precision | 61,5% | 63,6% | 76,2% | 0,0% | 77,8% | 40,0% | 62,5% |
| recall | 80,0% | 77,8% | 100,0% | / | 100,0% | 66,7% | 88,9% |
| f1 score | 69,6% | 70,0% | 86,5% | / | 87,5% | 50,0% | 73,4% |
| *focus on negative detection* | | | | | | | |
| precision | 95,6% | 94,3% | 100,0% | 100,0% | 100,0% | 97,4% | 97,2% |
| recall | 89,6% | 89,2% | 75,0% | 58,3% | 94,6% | 92,7% | 87,7% |
| f1 score | 92,5% | 91,7% | 85,7% | 73,7% | 97,2% | 95,0% | 92,2% |
| **with consideration of low confidence** | | | | | | | |
| accuracy | 90,9% | 88,7% | 89,1% | 73,7% | 96,2% | 93,0% | 90,5% |
| *focus on positive detection* | | | | | | | |
| precision | 68,8% | 63,6% | 76,2% | 0,0% | 77,8% | 50,0% | 64,7% |
| recall | 84,6% | 77,8% | 100,0% | / | 100,0% | 75,0% | 89,8% |
| f1 score | 75,9% | 70,0% | 86,5% | / | 87,5% | 60,0% | 75,2% |
| *focus on negative detection* | | | | | | | |
| precision | 96,7% | 95,2% | 100,0% | 100,0% | 100,0% | 98,0% | 97,9% |
| recall | 92,2% | 90,9% | 83,3% | 73,7% | 95,6% | 94,3% | 90,6% |
| f1 score | 94,4% | 93,0% | 90,9% | 84,8% | 97,7% | 96,2% | 94,1% |

| | |
|---|---|
| f1 score average without consideration of low confidence detections: | 82,8% |
| f1 score average with consideration of low confidence detections: | 84,7% |

Fig. 4. Results overview of main test drives.

## B. Low Confidence Switch Parameter Study

The impact of the method introduced in section Low Confidence Switch on classification performance and human time effort is analyzed by a parameter study varying the low confidence switch value between 50 % and 100 %. The result is shown in Figure 5.

The y-axis shows the F1 score average, and the x-axis shows the human time effort necessary to label the increasing detections flagged as low confidence. This time effort is calculated based on an average time consumption to label a detection. This varies based on engineering experience and detection complexity. For each of the six test drives of the main result data set the labeling time was measured and an average time per detection is calculated. The conservative (maximum) value of 4.67 seconds per detection is used in this diagram. Finally, the time consumption necessary is divided by the total time of the test drive. All six test drives sum up to a test drive of 26.3 minutes, which would equal 100 %. The analysis of the main result data set showed that a fully automated (low confidence value of 0 %) and therefore very fast (in terms of human effort) analysis achieves an F1 score average of 79.2 %. A low confidence switch of 80 % leads to an F1 score average of 90 % with 29.3 % time effort (equals 7.7 minutes for the main result data set) and an F1 score average of 100 % requires 89.97 % human time effort. Furthermore, the time effort for the analysis process used prior to this thesis is shown in red on a different scale. The value of 4000 % results of an analysis of a test drive lasting almost 30 minutes, and therefore comparable in scale with the main result data set, which took more than 20 hours to analyze (factor 40).

A comparison of the prior and new developed process on the same test drive data is not possible as both methods require a different camera setup in the vehicle. Considering the 100 % accuracy a pure theoretical value due to human errors, as pointed out at the end of the previous paragraph, the orange data point shows the approximation of the human error of 96.9 %, calculated based on five labeling errors occurred during the initial evaluation of main test drive data set. Accepting this classification performance as maximum achievable, the GTA-tool in its current implementation can reduce the workload by more than 40 %. The data shown only considers the human time effort to manually label the detections flagged as low confidence. The additional time consumption for starting the GTA-tool and inserting the path to the measurement file is neglected. A very conservative estimation of five minutes for this task is assumed to calculate the overall time savings.

The computation time is neglected as the GTA-tool could run on a server and can be accelerated by choosing the desired performance as it allows parallel computing. As a rule of thumb, the data preparation tool runs over night, the CNN classification takes approximately the same time as the analyzed test drive and the automation tool and decision maker run in a second.

For full manual mode the total human time effort is reduced from above 20 hours to approximately 27 min (90 % of test drive duration) by 97.75 %. For full automation mode, the total human time effort is reduced from above 20 hours to five minutes by 99.58 %.

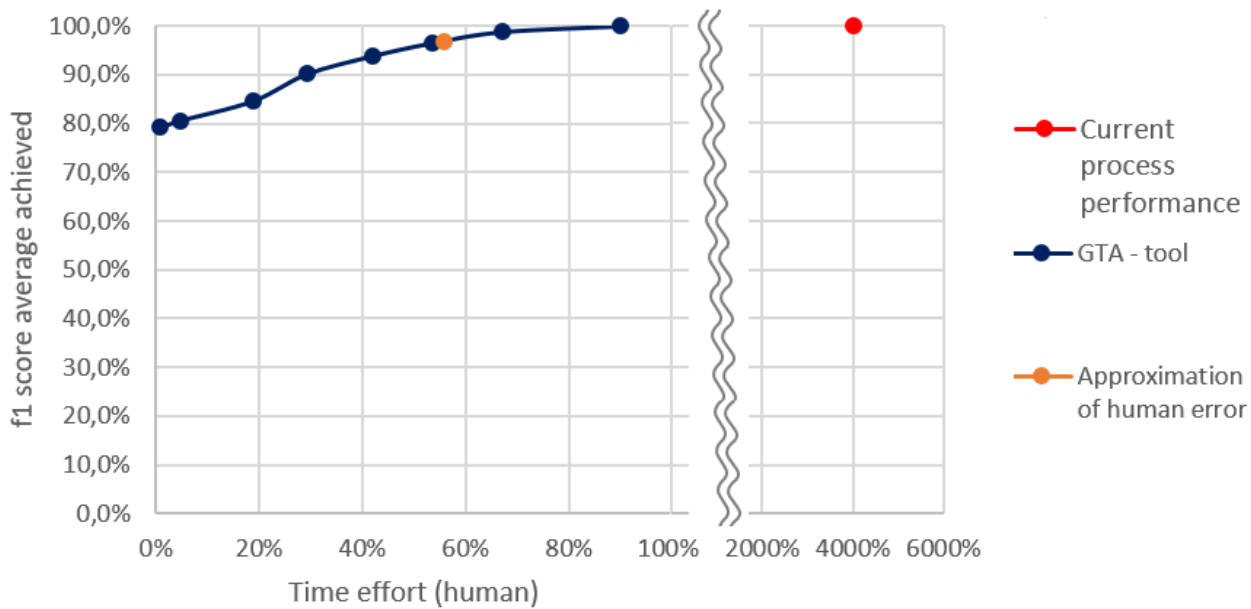

Fig. 5. Classification quality versus human time effort for analysis.

## V. Discussion

A concept to improve the current analysis process by using machine learning methods is presented and a tool using a preexisting CNN is implemented. Metrics are introduced to evaluate the classification performance of the GTA-tool. Findings based on the analysis of an initial data set with 77 detections lead to an optimized second version with focus on improved classification performance. Version 2 includes a low confidence switch, which flags detections with a weak CNN classification for human post labeling. The overall classification performance improved for the initial data set with the second version of the GTA-tool. Six ground truth test drives with a total of 304 detections are conducted providing additional data allowing for a statistically more relevant analysis of the GTA-tool performance.

The evaluation of the increased ground truth data set showed a total classification performance of 84.7 % with a low confidence switch value of 70 %. A deeper analysis of the false positive classifications unveiled four methods for further improvement, whereby more than 50 % of the false positives can be avoided with a road sign recognition method. Aiming for the highest classification performance achievable with the second version of the GTA-tool implemented, the impact of the low confidence switch is analyzed in a parameter study, showing the potential to reach a classification performance of up to 98.8 %. Higher performance comes at the cost of increased human time effort. This characteristic thus allows the user to select between full automation with a classification performance of 79.2 %, full manual assessment with potentially 100 % accuracy.

Although full automation with flawless classification performance has not yet been achieved yet, significant improvements compared to the previous analysis process are made. Even without using machine learning methods by CNNs, the human workload for a full manual assessment is reduced by a factor of 40 through user focused optimization of the video display and built-in labeling functions in the analysis sub module. Further on, test reports with main KPIs like false positive and F1 score are created automatically. A quantitative measure of the achievable classification performance by human test engineers is unlikely to be 100 %. Considering the test drives analyzed in this thesis, the relabeling of five cases reveals a realistic value of 96.9 %. Accepting the failure rate, a significant acceleration of test analysis can be achieved even with the upgraded version currently implemented. Further acceleration can be achieved with low effort as the analysis showed high potential for significant classification improvement with simple implementations as example given road recognition methods are already available in industry and also as open source.

## VI. Conclusions

The aim of this work is to reduce the work scope and time required for ground truth test analysis, necessary for the measure of main KPIs of a parking service. Within one iterative development optimization loop, considering the findings based on an initial test data set available, an automation tool is implemented, tested and available for test analysis in development projects.

Adjusting an internal parameter, the tool can be used for either full automation or maximum accuracy. The trade-off is evaluated and ranges from theoretically 100 % classification performance for full manual assessment via 96.6 % for 40 % automation to 79.2 % classification performance for full automation. The human time effort compared to the previous process is reduced by a factor of 40 (97.75 %) for full manual assessment, a factor of 100 for 40 % automation and takes only a few seconds for full automation mode. Further research could evaluate the use of automated drone video data sets [14], [15] for the creation of additional ground truth data sets.